# SWARM INTELLIGENCE

Sabu m Thampi, LBSITW {smtlbs@gmail.com}

*Biologically inspired computing is an area of computer science which uses the advantageous properties of biological systems. It is the amalgamation of computational intelligence and collective intelligence. Biologically inspired mechanisms have already proved successful in achieving major advances in a wide range of problems in computing and communication systems. The consortium of bio-inspired computing are artificial neural networks, evolutionary algorithms, swarm intelligence, artificial immune systems, fractal geometry, DNA computing and quantum computing, etc. This article gives an introduction to swarm intelligence.*

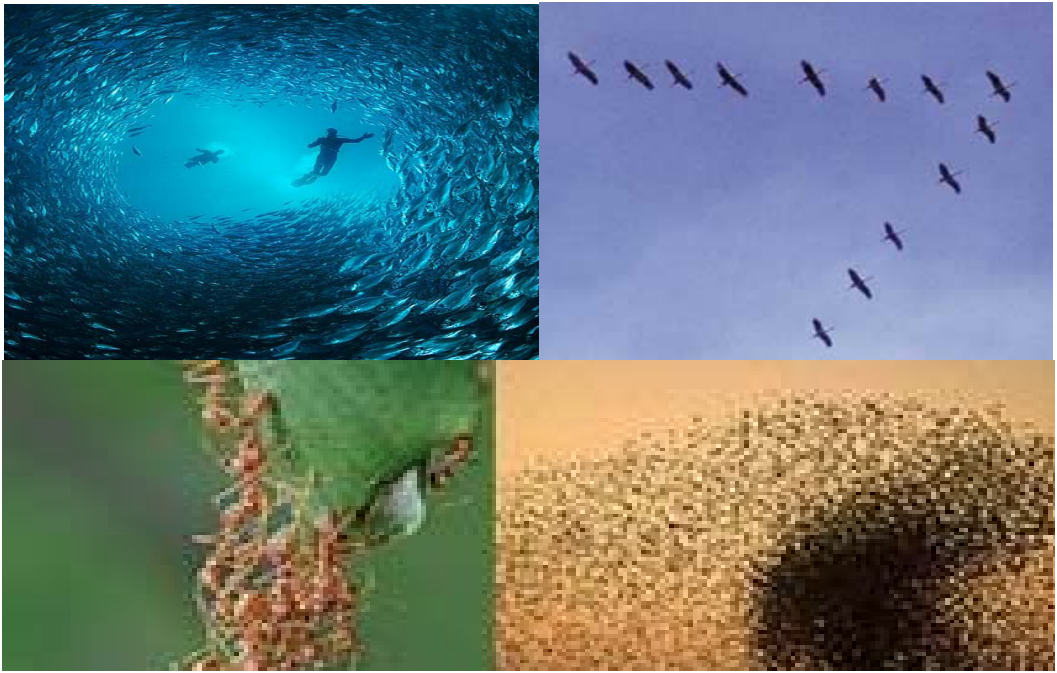

**A** long time ago, people discovered the variety of the interesting insect or animal behaviours in the nature. A congregate of birds sweeps across the sky. A group of ants hunt for food. A school of fish swims, turns, flees together, etc... We call this kind of aggregate motion *swarm behaviour*. Recently biologists and computer scientists in the field of *artificial life* (also called ALife - describes research into human-made systems that possess some of the essential properties of life) have studied how to model biological swarms to understand how such social animals interact, achieve goals, and evolve. Furthermore, engineers are more and more paying attention in this kind of swarm behaviour since the resulting "swarm intelligence" can be applied in optimization (e.g. in telecommunicate systems), robotics, traffic patterns in transportation systems, and military applications. In fact, there are already lots of computational techniques inspired by biological systems. For instance, artificial neural network (ANN) is a simplified model of human brain; genetic algorithm is inspired by the human evolution.

A high-level view of a swarm suggests that the N agents in the swarm are cooperating to achieve some purposeful behaviour and accomplish a few goals. This perceptible collective intelligence seems to materialize from what are often large groups of relatively simple agents. The agents use simple local rules to govern their actions and via the interactions of the entire group, the swarm achieves its objectives.

*Swarm intelligence* is the emergent collective intelligence of groups of simple autonomous agents. Here, an autonomous agent is a subsystem that interacts with its environment, which most likely consists of other agents, but acts relatively independently from all other agents. The autonomous agent does not follow commands from a leader. For example, for a bird to participate in a flock, it only adjusts its movements to coordinate with the movements of its flock mates, naturally its neighbours that are close to it in the flock. A bird in a flock merely tries to stay close to its neighbours, but evade collisions with them. Each bird does not take commands from any leader bird since there is no lead bird. Any bird can fly in the front, center and back of the swarm. Swarm behaviour helps birds take advantage of several things including protection from predators, and searching for food.

Swarm robotics is currently one of the most important application areas for swarm intelligence. Swarms provide the possibility of enhanced task performance, fault tolerance, less complexity and decreased cost over traditional robotic systems. They can accomplish some tasks that would be unfeasible for a single robot to attain. Swarm robots can be applied to many fields, such as flexible manufacturing systems, spacecraft, inspection/maintenance, construction, agriculture, and medicine work. Swarm robots are potentially reconfigurable networks of communicating agents capable of coordinated sensing and interaction with the environment.

There are two popular swarm inspired methods in computational intelligence areas: *ant colony optimization (ACO) and particle swarm optimization (PSO)*. The ACO introduced by Marco Dorigo in 1992 is one of the most successful techniques of the wider field of swarm intelligence. It is a probabilistic technique for solving computational problems which can be reduced to finding good paths through graphs. They are inspired by the behaviour of ants in finding paths from the colony to food. ACO has many successful applications in discrete optimization problems such as travelling salesman problem.

In the real world, ants (initially) wander randomly, and upon finding food return to their colony while laying down pheromone trails. If other ants find such a path, they are likely not to keep travelling at random, but to instead follow the trail, returning and reinforcing it if they eventually find food. Over time, however, the pheromone trail starts to evaporate, thus reducing its attractive strength. The more time it takes for an ant to travel down the path and back again, more time the pheromones have to evaporate. A short path, by comparison, gets marched over faster, and thus the pheromone density remains high as it is laid on the path as fast as it can evaporate. Pheromone evaporation has also the advantage of avoiding the convergence to a locally optimal solution. If there were no evaporation at all, the paths chosen by the first ants would tend to be excessively attractive to the following ones. In that case, the exploration of the solution space would be constrained. Thus, when one ant finds a good path from the colony to a food source, other ants are more likely to follow that path, and positive feedback eventually leads all the ants following a single path.

The idea of the ant colony algorithm is to mimic the ant behaviour with "simulated ants" walking around the graph representing the problem to solve. Assuming that the problem to be optimized can be represented by a graph, the general ACO algorithm can be described as given below:

1. Initialization: assign the same initial pheromone value to each edge of the graph, and randomly place an ant in a location of the search space.

2. Population loop: for each ant, do:

   a. *Probabilistic transition rule:* according to a given probabilistic transition rule, move ant over the space so that a solution to the problem is built.

   b. *Goodness evaluation:* evaluate the goodness of the solution obtained by this ant.

   c. *Pheromone updating:* update the pheromone level of each edge by reinforcing good solutions. Reduce the pheromone level of each edge (evaporation).

3. Repeat Step 2 until a given convergence criterion is met.

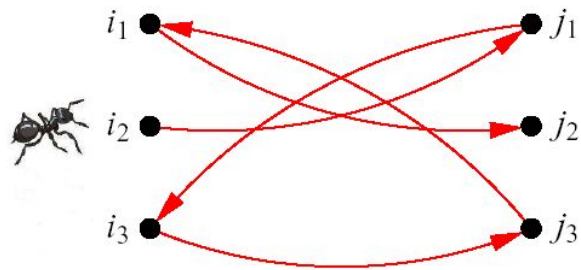

**P**article Swarm Optimization method is motivated from the simulation of social behaviour of bird flocking and fish schooling. PSO was originally designed and developed by Eberhart and Kennedy. However, it shares many similarities with evolutionary computation techniques such as genetic algorithms (GA). The system is initialized with a population of random solutions and searches for optima by updating generations. Unlike GA, PSO has no evolution operators such as crossover and mutation. In PSO, the potential solutions fly through the problem space by following the current optimum. Each single solution is a "bird" in the search space and it is called as a "particle". All of particles have fitness values which are evaluated by the fitness function to be optimized and have velocities which direct the flying of the particles. The particles fly through the problem space by following the current optimum particle called *guide*.

PSO is initialized with a group of random particles (solutions) and then searches for optima by updating generations. In each iteration, every particle is updated by following two "best" values. The first one is the best solution (fitness) it has achieved so far. This value is called *pbest*. Another "best" value that is tracked by the particle swarm optimizer is the best value, obtained so far by any particle in the population. This best value is a global best and called *gbest*. When a particle takes part of the population as its topological neighbours, the best value is a local best and is called *lbest*.

After finding the two best values, the particle updates its velocity and positions with following equation (a) and (b).

$$v[] = v[] + c_1 * rand() * (pbest[] - present[]) + c_2 * rand() * (gbest[] - present[]) \quad\ldots\ldots(a)$$

$$present[] = present[] + v[] \quad\ldots\ldots\ldots\ldots\ldots\ldots(b)$$

where, v[] is the particle velocity, present[] is the current particle (solution). rand() is a random number between (0,1). $c_1$, $c_2$ are learning factors. Usually $c_1 = c_2 = 2$. Particles' velocities on each dimension are clamped to a maximum velocity $V_{max}$.

*The pseudo code of the procedure is as follows:*

```
For each particle
    Initialize particle
End
Do
   For each particle
      Calculate fitness value
      If the fitness value is better than the best fitness value
        (pbest) in history
          Set current value as the new pbest
   End
   Choose the particle with the best fitness value of all the
     particles as the gbest
   For each particle
      Calculate particle velocity according equation (a)
      Update particle position according to equation (b)
   End
While maximum iterations or minimum error criteria is not
attained
```

Ant colony optimization and particle swarm optimization techniques utilise extremely simple algorithms that seem to be effective for optimizing a wide range of functions. Since the introduction of the algorithms, significant contributions on algorithmic variants, challenging application problems, and theoretical foundations have been obtained. These have established swarm intelligence techniques as a high-level algorithmic framework for the solution of difficult optimization problems.